\documentclass{article}
\usepackage{spconf,amsmath,graphicx}

\usepackage{color}

\title{Topic-Aware Pointer-Generator Networks for\\ Summarizing Spoken Conversations}

\name{Zhengyuan Liu$^{\dagger}$, Angela Ng$^{\ddagger}$, Sheldon Lee$^{\ddagger}$, Ai Ti Aw$^{\dagger}$, Nancy F. Chen$^{\dagger}$\sthanks{This research was supported by funding for Digital Health from the Institute for Infocomm Research (I2R) and the Science and Engineering Research Council (Project No. A1818g0044), A*STAR. This work was conducted using resources from the Human Language Technology unit at I2R. We thank R. E. Banchs, L. F. D'Haro, P. Krishnaswamy, H. Lim, F. A. Suhaimi and S. Ramasamy at I2R, and W. L. Chow, A. Ng, H. C. Oh, and S. C. Tong at Changi General Hospital for insightful discussions.}}

\address{$^{\dagger}$Institute for Infocomm Research, A*STAR, Singapore\\$^{\ddagger}$Changi General Hospital, Singapore}

\begin{document}
%
\maketitle
\begin{abstract}
Due to the lack of publicly available resources, conversation summarization has received far less attention than text summarization.
As the purpose of conversations is to exchange information between at least two interlocutors, key information about a certain topic is often scattered and spanned across multiple utterances and turns from different speakers.
This phenomenon is more pronounced during spoken conversations, where speech characteristics such as backchanneling and false-starts might interrupt the topical flow. Moreover, topic diffusion and (intra-utterance) topic drift are also more common in human-to-human conversations.
Such linguistic characteristics of dialogue topics make sentence-level extractive summarization approaches used in spoken documents ill-suited for summarizing conversations. 
Pointer-generator networks have effectively demonstrated its strength at integrating extractive and abstractive capabilities through neural modeling in text summarization. To the best of our knowledge, to date no one has adopted it for summarizing conversations.
In this work, we propose a topic-aware architecture to exploit the inherent hierarchical structure in conversations to further adapt the pointer-generator model. Our approach significantly outperforms competitive baselines, achieves more efficient learning outcomes, and attains more robust performance.
\end{abstract}

\section{Introduction}
\label{sec:intro}
Automatic summarization condenses lengthy materials into shorter versions, which focuses on the essential information and the overall meaning.
Such summaries can enable users to browse and digest information more effectively and efficiently, which is especially useful in the digital era of information overload. Thus, summarization has attracted much research attention over the years.
When humans generate summaries, they usually first comprehend the entire content, and then extract and concatenate the keywords or salient sentences. To obtain more succinct and readable results, they further distill and rephrase the most important information \cite{jing-1999-decomposition}. This gives rise to the two main paradigms in automatic summarization: extractive and abstractive.
Recently, with sophisticated neural architectures, representation learning of linguistic elements and large-scale available corpora, data-driven approaches have made much progress in both two paradigms \cite{nallapati-2017-summarunner, nallapati-2016-abs-beyond, Liu-extractive-spoken-Sum}.
Most work in text summarization focuses on documents such as news or academic articles \cite{hermann-2015-CnnDaily-Data, elhadad-2005-medical-summary}.
On the contrary, due to the scarcity of publicly available corpora with ground-truth summaries, the speech summarization task of human-to-human spoken conversations has received far less attention even though there is high industry demand and many potential applications across different domains \cite{zhang2007-lecture-Sum,goo-2018-meeting-summary}.

Different from passages, human-to-human spoken conversations are a dynamic flow of information exchange, which is often informal, verbose and repetitive, sprinkled with false-starts, backchanneling, reconfirmations, hesitations, and speaker interruptions \cite{sacks-1978-turn-taking}. The key information about a certain topic is often at the sub-sentence or sub-utterance level, and scattered and spanned across multiple utterances and turns from different speakers, leading to lower information density, more diffuse topic coverage, and (intra-utterance) topic drifts. These spoken characteristics pose technical challenges for sentence-level extractive approaches that would inevitably include unnecessary spans of words in the generated summaries. Pointer-generator networks \cite{see-2017-PGnet}, a neural sequence-to-sequence design \cite{sutskever-2014-seq2seq-paper}, produce summaries via word-level extraction and abstractive generation.
Thus, we propose to exploit its advantages to tackle the aforementioned challenges. Meanwhile, although conversations are often less structured than passages, they are inherently organized around the dialogue topics in a coarse-grained structure \cite{sacks-1978-turn-taking}. Topic segmentation has also been shown to be useful in dialogue-based information retrieval \cite{boufaden-2001topic-segment}.
Such prior analyses and investigations inspire us to augment pointer-generator networks with a topic-level attention mechanism to more elegantly attend to the underlying (yet often interrupted) topical flow in human-to-human conversations. 

The dialogue setting in this work is based on a real-world scenario where nurses discuss symptom information with post-discharge patients to follow-up with their status. The conversation summarization task here is targeted at automatically generating notes that describe the symptoms the patients are experiencing. Our proposed topic-aware pointer-generator framework exploits the inherent hierarchical structure in dialogues and is able to address the technical challenges of spoken dialogue summarization: empirical results outperform competitive baseline architectures significantly, while achieving more efficient and robust learning outcomes. 

\section{Related Work}
\label{sec:related_work}
In speech and text summarization, traditional approaches are studied more on the extractive methods, utilizing rule-based, statistical \cite{hong-2014-rankSum} and graph-based \cite{Erkan-2004-LexRank} algorithms with various linguistic features like lexical similarity \cite{saggion-2013-past-sum}, semantic structure \cite{zhang2007-lecture-Sum}, or discourse relation \cite{Hirao-2015-TrimTreeSum}. 
Recently, end-to-end neural approaches have been widely adopted in text summarization due to their capability, flexibility and scalability. Neural extractive models adopt sentence labeling or ranking strategies \cite{kedzie-2018-contentSelection,narayan-etal-2018-rankingSum}, semantic vector representations \cite{pennington-2014-glove} and sequential context modeling \cite{hochreiter1997-LSTM}. For abstractive tasks, sequence-to-sequence models use a neural decoder to generate informative and readable summaries word-by-word \cite{nallapati-2016-abs-beyond}. Various methods have been proposed for improvement: the attention mechanism helps the decoder concentrate on appropriate parts of source content \cite{chopra-2016-attentive-seq2seq}; the pointer mechanism is effective in handling out-of-vocabulary words \cite{see-2017-PGnet}; the coverage mechanism is used to reduce generative repetitions \cite{paulus-2017-RL-Sum}. Extractive-abstractive models have also been proposed to obtain better results \cite{hsu-2018-inconsistent-Sum, gehrmann-2018-bottom-up}.

While text summarization focuses on passages such as news articles \cite{hermann-2015-CnnDaily-Data, rush-2015-neural-abs} and academic publications \cite{elhadad-2005-medical-summary}, speech summarization has been investigated in monologues (such as broadcasts \cite{Hirschberg-2005-speech-sum} and lectures \cite{zhang2007-lecture-Sum}) and multi-party dialogues such as meetings \cite{liu-yang-2010-meeting-sum}. Recently, neural modeling approaches have also been adopted: Goo \textit{et al.} \cite{goo-2018-meeting-summary} used a sequence-to-sequence model to write headlines for meetings, and Liu \textit{et al.} \cite{Liu-extractive-spoken-Sum} used a hierarchical extractive model to summarize monologue spoken documents. In this paper, we propose a topic-aware pointer-generator architecture for hierarchical context modeling to generate summary notes of spoken conversations.

\section{Conversation Corpus and Setup}
\label{sec:corpus}
We sampled 100k dialogues as the training set and another distinct 1k dialogues as the validation set according to Section \ref{ssec:dialogue_data}. While the training and validation sets were constructed using simulated data, the test set was derived from 490 multi-turn dialogues that took place between nurses and patients in the healthcare setting. Topic segmentation (Section \ref{ssec:segmentation}) and ground-truth summary construction (Section \ref{ssec:groud-truth}) were conducted on all these subsets.

\subsection{Nurse-to-Patient Dialogue Data}
\label{ssec:dialogue_data}
This corpus was inspired by a pilot set of conversations that took place in the clinical setting where nurses inquire about symptoms of patients \cite{liu-etal-2019-fast}. Linguistic structures at the semantic, syntactic, discourse and pragmatic levels were systemically abstracted from these conversations to construct templates for automatically simulating multi-turn dialogues \cite{shah-2018-dialogue-simulator}. The informal and spontaneous styles of spoken interactions such as interlocutor interruption, backchanneling, hesitation, false-starts, repetition, and topic drift were preserved (see Figure \ref{fig:dialogue_example}a for an example).

A team of linguistically trained personnel refined, substantiated, and corrected the simulated dialogues by enriching verbal expressions through considering paraphrasing, different regional English speaking styles (American, British, and Asian) through word usage and sentence patterns, validating logical correctness through considering if the dialogues were natural and not disobeying common sense, and verifying the clinical content by consulting certified nurses. These conversations cover 9 topics/symptoms (e.g., headache, cough).

\subsection{Topic Segmentation}
\label{ssec:segmentation}
In dialogue analysis, a change in topic corresponds to a change in cognitive attention acknowledged and acted upon by speakers, which is usually related to content themes \cite{boufaden-2001topic-segment}. In this work, we specify the dialogue topics according to the symptoms in nurse-to-patient conversations. Figure \ref{fig:dialogue_example}a shows an example of different topic segments, where each spans across different utterances and speakers. Note that various types of spoken characteristics (e.g. false start) could break up topical congruence. 

A rule-based lexical algorithm was used to detect the boundaries between dialogue topics. Labels $\langle s \rangle$ and $\langle /s \rangle$ were respectively added before and after each topic segment.
Position indices of segment labels are used in Section \ref{ssec:proposed_model}. Human verification was conducted to ensure quality control.

\subsection{Ground-Truth Summaries}
\label{ssec:groud-truth}
The goal of this conversation summarization task is to obtain a concise description characterizing different attributes of a specified symptom. 
For this particular clinical scenario, the summary notes are preferred to be represented in a very structured format to facilitate indexing, searching, retrieving and extracting in a variety of downstream workflow applications (e.g., decision support, medical triage). Thus, paraphrases of a particular symptom are represented using the same entity, e.g., \textit{"shortness of breath"} and \textit{"breathlessness"} are both represented as symptom \textit{"breathlessness"}.

\begin{figure}[t]
\centering
\centerline{\includegraphics[width=7cm]{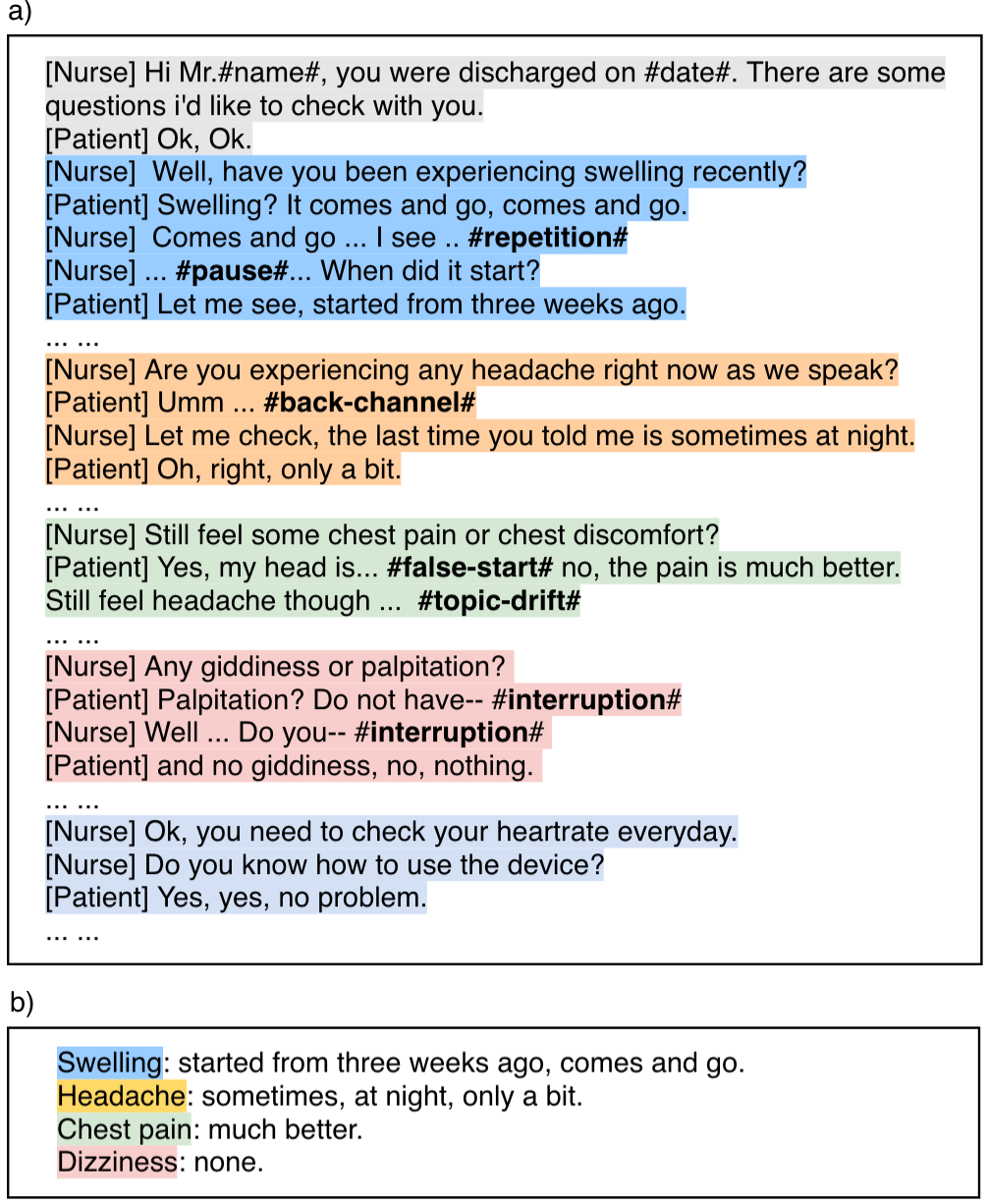}}
\caption{a) A dialogue example of multi-turn conversation on nurse-patient symptom discussion. Colored spans are utterances of the same topic. Spoken characteristics are preserved and represented in bold font; b) Generated ground-truth summary. Colored spans indicate corresponding topics in the given dialogue example.}
\vspace{-0.3cm}
\label{fig:dialogue_example}
\end{figure}

The summary format is shown in Figure \ref{fig:dialogue_example}b, where each symptom is listed separately with corresponding attributes such as frequency of symptom or severity of symptom that were mentioned in the conversations. If a symptom was mentioned, it will be included in the summary. If there is no signal of a symptom (e.g., \textit{``cough''}) in the discussion between the nurse and the patient, the summary for the symptom is represented as \textit{``Cough: none''}, while the others would be recorded with key information from the dialogue, e.g., \textit{``Headache: every night, only a bit''}. Human verification was conducted to ensure quality control.

\section{Approach}
\label{sec:models}
In a sequence-to-sequence (seq2seq) model, the encoder receives a token sequence of content $x=\{x_0, x_1, ... , x_n\}$ of length $n$, and the decoder outputs a token sequence of summary $y=\{y_0, y_1, ..., y_m\}$ of length $m$. The decoder generates words step by step, and previously generated words are encoded to provide contextual information at each time step. The task is to learn a function $f$ with a parameter set $\theta$ that maximizes the probability to generate readable and meaningful output text. In this section, we (1) describe two baselines: an attentive seq2seq model and a pointer-generator network, and (2) demonstrate how we integrate topic-level attention neural mechanisms.

\subsection{Attentive Seq2Seq Architecture}
\label{ssec:seq2seq}
The attentive seq2seq model is similar to that in \cite{chopra-2016-attentive-seq2seq}. It adds an attention layer to a vanilla seq2seq network, to filter out unnecessary contextual information at each decoding step.

\noindent{\textbf{Sequence Encoding}}: Given a document input $x$ (one-hot word representation), an embedding layer converts it to vector representations by a look-up operation using the embedding matrix, obtaining a vector sequence $v=\{v_0, v_1, ... , v_n\}$, where $v_i \in R^{d}$ and $d$ is the embedding dimension. Then, a bi-directional long short-term memory (Bi-LSTM) layer \cite{schuster-1997-biLSTM} is used to encode $v_i$ with forward and backward temporal dependencies, as $\overrightarrow{h_i}$ and $\overleftarrow{h_i}$, respectively, and concatenate them as the hidden representation $h=\{h_0, h_1, ... , h_n\} \in R^{n\times 2d^{'}}$, where $d^{'}$ is the hidden dimension size.
\begin{equation}
\overrightarrow{h_{i}}=\overrightarrow{\mbox{LSTM}}(v_i, \overrightarrow{h_{i-1}});\ \ \overleftarrow{h_{i}}=\overleftarrow{\mbox{LSTM}}(v_i, \overleftarrow{h_{i+1}})
\end{equation}
\begin{equation}
h_i=[\overrightarrow{h_{i}};\overleftarrow{h_{i}}]
\end{equation}

\noindent{\textbf{Sequence Decoding with Attention}}: The decoder is a single-layer unidirectional LSTM, generating words step by step. For each decoding step $t$, it receives the word embedding of the previous token $y_{t-1}$, then calculates the decoder state $s_t$. Attention scoring is conducted through concatenation \cite{luong-2015-attention}:
\begin{equation}
\label{eq:pointer_dis}
a^t = \mbox{softmax}(W_e\mbox{tanh}(W_{attn}[h_{i};s_{t}]+b_{attn}))
\end{equation}
where $W_e$, $W_{attn}$ and $b_{attn}$ are trainable parameters, and $[;]$ is the concatenation operation. Attention scores can be viewed as the importance over the input content, guiding the decoder to concentrate on the appropriate positions of context for generating the next word. Next, these attention scores are used to produce a weighted-pooling of the encoded hidden states, as a fixed-size representation of what has been read from the source for this step, namely the context vector $h_t^c$.

Then, $h_t^c$ is concatenated with the decoder state $s_t$ and fed through two dense layers to produce a distribution on the vocabulary:
\begin{equation}
\label{eq:vocab_dis}
p^t_{vocab} = \mbox{softmax}(W^{''}(W^{'}[s_t;h_t^{c}]+b^{'})+b^{''})
\end{equation}
where $W^{'}$, $W^{''}$, $b^{'}$ and $b^{''}$ are trainable parameters. $p^t_{vocab}$ is a probability distribution over all words in the vocabulary, which will be used to generate decoded tokens.

\begin{figure*}[t]
\centering
\centerline{\includegraphics[width=15cm]{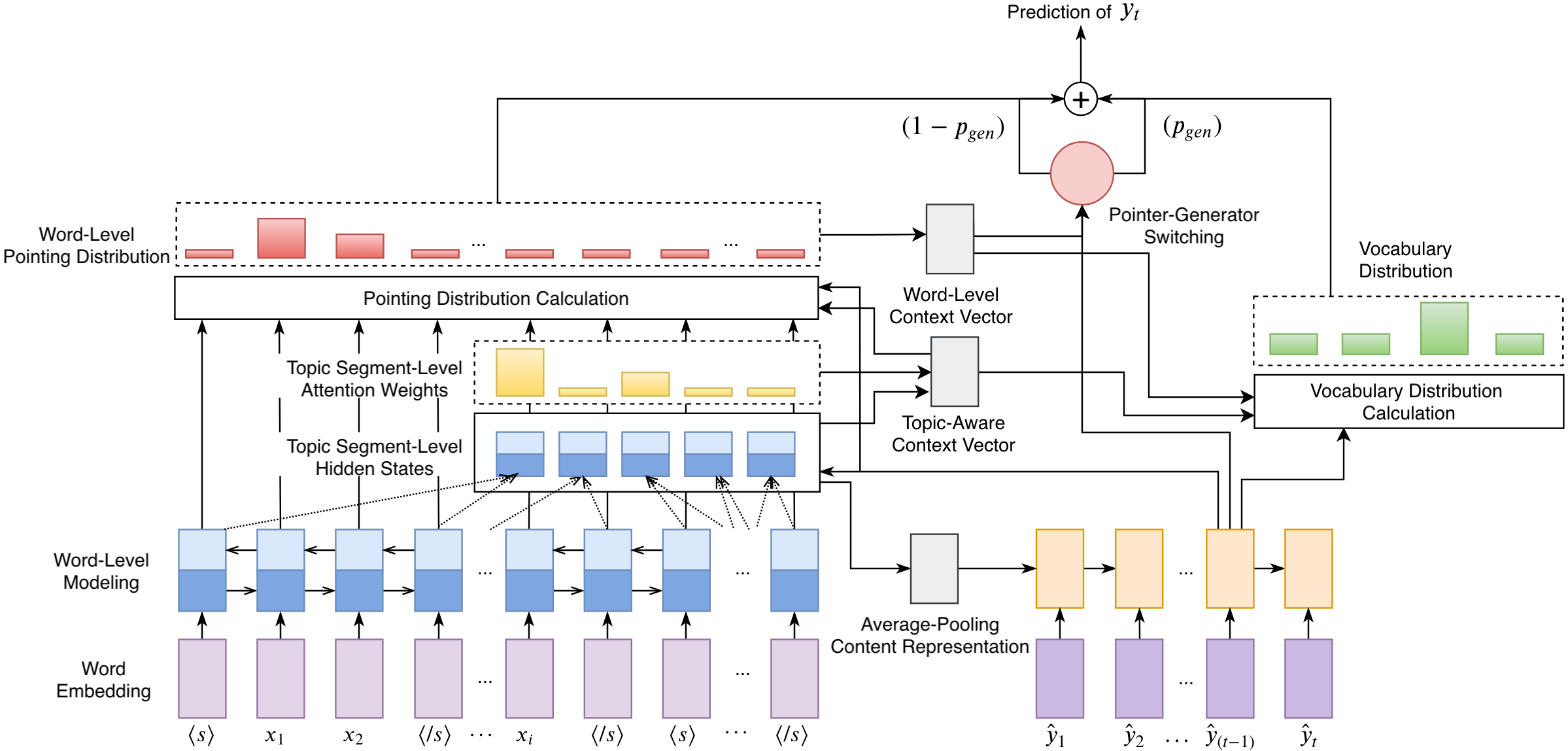}}
\vspace{-0.2cm}
\caption{The architecture of our proposed topic-aware pointer-generator network.}
\label{fig:model_architecture}
\vspace{-0.2cm}
\end{figure*}

\subsection{Pointer-Generator Networks}
\label{ssec:pg_net}
Pointer-generator networks are a variant of seq2seq architecture by adding a pointer network \cite{NIPS2015-pointNet}. Aside from generating words over a fixed vocabulary, the model is able to copy words via pointing to the source content, thus bypassing out-of-vocabulary issues. In the pointer-generator model, the sequence encoding representation $h$, attention distribution $a^t$ and context vector $h_t^{c}$ are calculated as in Section \ref{ssec:seq2seq}. We describe how the pointer-generator model achieves word-level extraction and abstractive generation below. 

\noindent{\textbf{Pointer Mechanism}}:
To directly copy words from the source content, the attentive distribution $a^t$ in Equation (\ref{eq:pointer_dis}) is regarded as the copy probability on input sequence $x$, specifically, the token on position with max probability will be extracted as the output.

\noindent{\textbf{Pointer-Generator Switching}}: At each decoding step, there is a switching probability to determine whether to generate a token from the fixed vocabulary or copy one from the source. $p_{gen} \in [0, 1]$ for step $t$ is calculated from the context vector $h^t_{c}$, the decoder input $y_{t-1}$ and the decoder state $s_t$:
\begin{equation}
\label{eq:p-gen}
p^t_{gen}=\sigma(W_{ptr}[h^t_c;y_{t-1};s_t]+b_{ptr})
\end{equation}
where $W_{ptr}$ and $b_{ptr}$ are trainable parameters and $\sigma$ is the sigmoid function. Then, $p^t_{gen}$ is used as a soft switch to choose between copying a word from the input sequence with attention distribution $a^t$, or generating a word from the vocabulary distribution $p^t_{vocab}$ in Equation (\ref{eq:vocab_dis}). For each sample, we extend the vocabulary with the unique words from the input content. We obtain the following probability distribution over the extended vocabulary:
\begin{equation}
\label{eq:pg_output}
p^t_{output}= p^t_{gen}\times p^t_{vocab}(w) + (1-p^t_{gen}) \times \sum_{i:w_i=w}a^t_i
\end{equation}
Contrary to the vanilla seq2seq model, which is restricted to a fixed vocabulary, the ability to copy and generate words is one of the primary advantages of the pointer-generator design. Another improvement over vanilla seq2seq is its coverage mechanism that avoids repetitions during decoding. More details can be found in \cite{see-2017-PGnet}.

\subsection{Topic-Aware Attentive Architecture}
\label{ssec:proposed_model}
To exploit topical structure in dialogue modeling with a hierarchical architecture, another attention layer is introduced to score topic-level attention to obtain a topic-aware context representation. We delineate how we integrate topic-aware attention to the baseline models below (see Figure \ref{fig:model_architecture}).

\noindent{\textbf{Topic-Level States}}: We obtain the representations $h^{seg}$ of topic segments by collecting hidden states from $h$ in equation (1) with the topic-level segment position indices $t^{seg}=\{t^{seg}_0,t^{seg}_1,...t^{seg}_k\}$, where $k$ is the topic segment number of the dialogue content. For the output of bi-directional LSTM, we collect the states of forward and backward directions and then concatenate them into one.

\noindent{\textbf{Initial Decode State}}: In the two baseline models, the last hidden state of sequence encoding representation is used as the initial state $s_0$ fed to the decoder. Here, we denote average-pooling of topic-level states $h^{seg}$ as $s_0$.

\begin{table*}[]
\center
\begin{tabular}{llllllllll}
\hline
\textbf{Model} & \multicolumn{3}{l}{\textbf{ROUGE-1}} & \multicolumn{3}{l}{\textbf{ROUGE-2}} & \multicolumn{3}{l}{\textbf{ROUGE-L}} \\ \hline
\textbf{} & \textbf{F1} & \textbf{Precision} & \textbf{Recall} & \textbf{F1} & \textbf{Precision} & \textbf{Recall} & \textbf{F1} & \textbf{Precision} & \textbf{Recall} \\ \hline
Attn Seq2Seq & 0.4494 & 0.3495 & 0.8312 & 0.3444 & 0.2664 & 0.6582 & 0.4074 & 0.3171 & 0.7542 \\ 
Attn Seq2Seq + TA & 0.5041 & 0.3922 & 0.8230 & 0.3874 & 0.2998 & 0.6491 & 0.4560 & 0.3595 & 0.7482 \\
PG-Net & 0.5345 & 0.4238 & 0.8653 & 0.4311 & 0.3341 & 0.7220 & 0.4949 & 0.3843 & 0.8159 \\ 
PG-Net + TA (Proposed) & \textbf{0.5862} & 0.4803 & 0.8703 & \textbf{0.4906} & 0.3928 & 0.7550 & \textbf{0.5496} & 0.4423 & 0.8260 \\ \hline
\end{tabular}
\caption{\label{result-table} Evaluation results of baselines and the proposed model.}
\end{table*}

\noindent{\textbf{Topic-Aware Contextual Representation}}: In each decoding time step $t$, we calculate topic-level attention and use topic-aware context vectors as the guide for the fine-grained word-level prediction.
First, the topic-level attention scores are calculated via dense layers and softmax normalization:
\begin{equation}
\label{eq:topic_attn}
a_t^{seg} = \mbox{softmax}(W_e^{seg}\mbox{tanh}(h_{i}W_{seg}s_{t}+b_{seg}))
\end{equation}

Next, we multiply the attention score with topic-level states to obtain the topic-aware context vector:
\begin{equation}
h_t^{*} = \sum_i^k h^{seg}a_t^{seg}
\end{equation}

Then, the pointing distribution in Equation (\ref{eq:pointer_dis}) and vocabulary distribution in Equation (\ref{eq:vocab_dis}) are influenced by the topic-aware contextual representation, and final output is produced as in Equation (\ref{eq:pg_output}):
\begin{equation}
a^t = \mbox{softmax}(W_e\mbox{tanh}(W_{attn}[h_{i};h_t^{*};s_{t}]+b_{attn}))
\end{equation}
\begin{equation}
p^t_{vocab} = \mbox{softmax}(W^{''}(W^{'}[s_t;h_t^{c};h_t^{*}]+b^{'})+b^{''})
\end{equation}

\section{Experiments}
\label{sec:experiment}

\subsection{Training Setup}
The experiments are conducted on the nurse-to-patient conversation corpus described in Section \ref{sec:corpus}. We implemented an attentive seq2seq (\textit{Attn Seq2Seq}) and a pointer-generator network (\textit{PG-Net}) as baselines, a topic-aware attentive seq2seq (\textit{Attn Seq2Seq+TA}) model as control, and our proposed topic-aware pointer-generator model (\textit{PG-Net+TA}).

Cross entropy is used to measure the loss between prediction and ground-truth. For time step $t$, the negative log likelihood of the target word $\hat{y_t}$ is defined as:
\begin{equation}
loss_t = -\mbox{log}P_{\theta}(\hat{y}_{t}|x)
\end{equation}
and the overall loss is the sum from all the time steps. Teacher forcing strategy is applied: during training, the input is the previous word from the ground truth; at test time, the input is the previous word predicted by the decoder. 

In our setup, for both the encoder and decoder, the dimension of word embeddings and hidden states was set to $200$. We adopted pre-trained word embedding Glove \cite{pennington-2014-glove}, and out-of-vocabulary words and segment labels were initiated with random vectors. Embedding weight sharing strategy was applied by sharing the same embedding matrix $W_{emb}$ for both encoder and decoder. This sharing significantly reduced parameter size and boosted the performance by reusing the semantic and syntactic information in one embedding space \cite{paulus-2017-RL-Sum}. The learning rate was fixed to $0.0001$ and the batch size was set to $32$. We adopted gradient clipping with a maximum gradient norm of $2.0$. Adam algorithm \cite{kingma2014adam} was used for stochastic optimization, with $\beta_1=0.9,\beta_2=0.99$. The vocabulary size was $10K$. We limited source contents to $300$ tokens and the decoding length to $100$ tokens. We adopted early-stop strategy with validation in each training epoch. During testing, we set the beam search size to $5$.

\begin{figure}[t]
\centering
\centerline{\includegraphics[width=5.2cm]{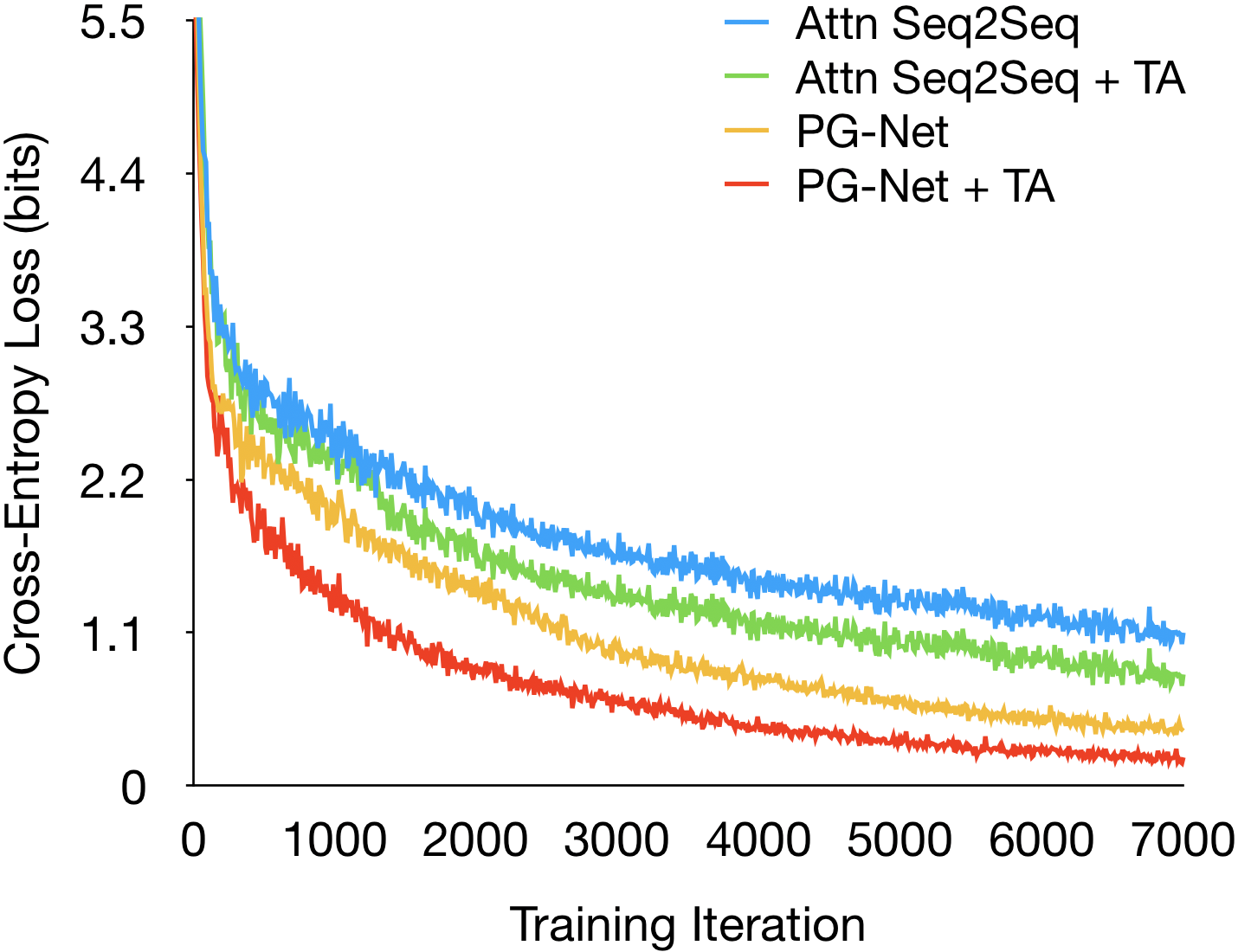}}
\vspace{-0.2cm}
\caption{Evaluation results of learning efficiency.}
\label{fig:loss_comparison}
\vspace{-0.2cm}
\end{figure}

\begin{figure*}[t]
\centering
\centerline{\includegraphics[width=16cm]{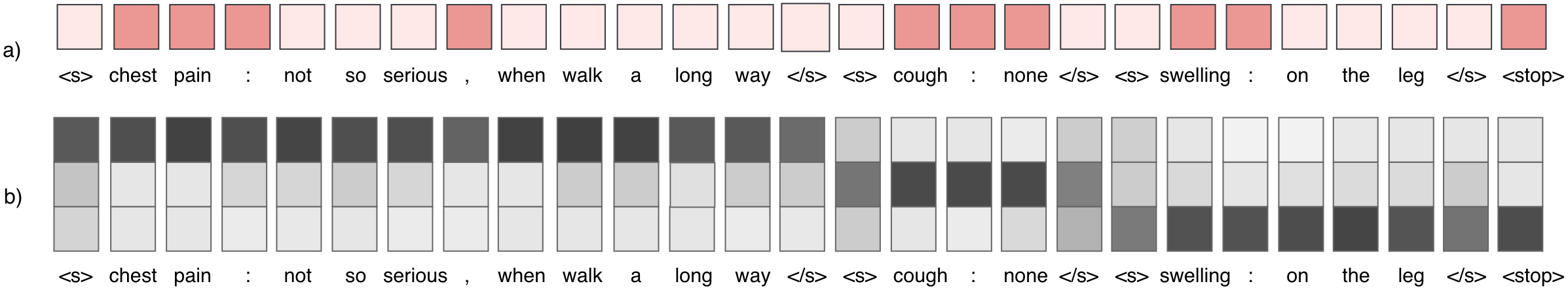}}
\caption{a) Switching between pointer \& generator: Tokens with higher switching probability are darker in color and are generated from the vocabulary. b) Visualizing topic-level attention scores. The darker the shade, the higher the scores.}
\label{fig:attn_visual}
\end{figure*}

\subsection{Empirical Evaluations}

\subsubsection{Evaluation I: Effectiveness}
\label{ssec:effectiveness_eval}
Summarization performance is measured using ROUGE-1, ROUGE-2 and ROUGE-L scores \cite{lin-2004-rouge}, as shown in Table \ref{result-table}. Between the two baselines, the pointer-generator model obtains higher performance than attentive seq2seq model. Qualitative analysis (Section \ref{sssec:switching_prob_vis}) shows that a certain proportion of generated tokens are directly copied from the source content, demonstrating the effectiveness of using the pointer mechanism in our task.
Moreover, with topic-aware attentive modeling, both baseline models obtain significant improvement, and the proposed \textit{PG-Net+TA} achieves the best performance. The gains are more prominent for the precision scores (see Table \ref{result-table}), indicating that the proposed approach generates fewer unnecessary tokens while preserving key information in the generated summary.
 
\subsubsection{Evaluation II: Learning Efficiency}
To evaluate the learning efficiency of the models, we recorded their loss values during training. As shown in Figure \ref{fig:loss_comparison}, in the first 7000 batches of iterations, loss of the pointer-generator decreases faster than that of attentive seq2seq. Moreover, by adding topic-level attention, both attentive seq2seq and pointer-generator are improved, and our proposed \textit{PG-Net+TA} performs best.\\

\noindent Having demonstrated the strength of pointer-generator networks over attentive seq2seq model, we will focus on the former in the following experiments.

\subsubsection{Evaluation III: Performance/Model Robustness}
Spoken conversations are often verbose with low information density scattered with topics not central to the main dialogue theme, especially since speakers chit-chat and get distracted during task-oriented discussions. To evaluate such scenarios, we adopted model-independent ADDSENT \cite{jia-liang-2017-adversarial}, where we randomly extracted sentences from SQuAD and inserted them before or after topically coherent segments. The average length of the augmented test set, is increased from $300$ to $900$. As shown in Table \ref{lengthy-table}, topic-level attention helps the pointer-generator model become more robust to lengthy samples.

\subsubsection{Evaluation IV: Low Resource Training}
\begin{figure}[t]
\centering
\centerline{\includegraphics[width=7.2cm]{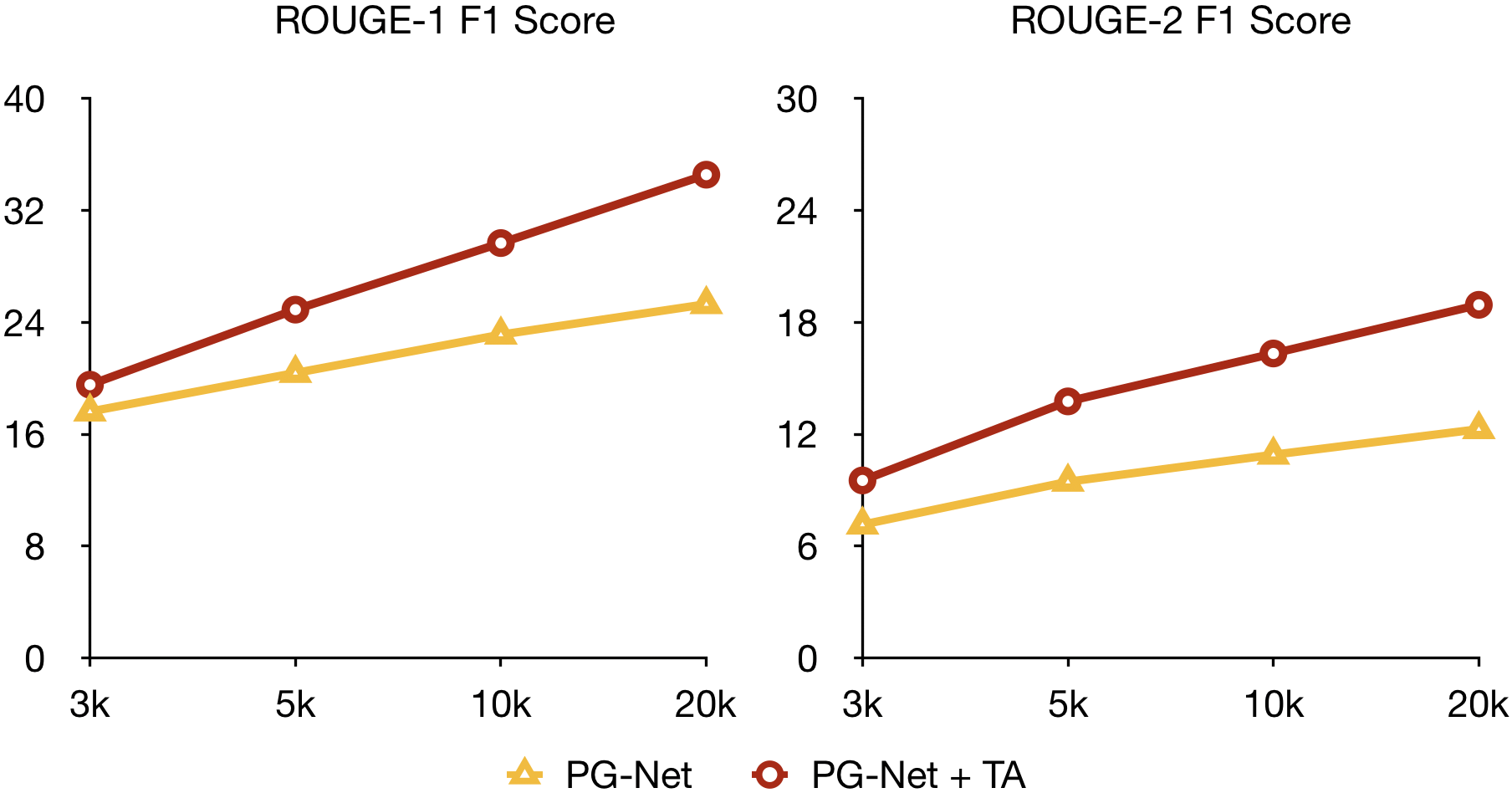}}
\vspace{-0.2cm}
\caption{Evaluation results of low resource training.}
\label{fig:low_resource}
\vspace{-0.2cm}
\end{figure}

Limited amount of training data is a major pain point for dialogue-based tasks, as it is time-consuming and labor-intensive to collect and annotate natural dialogues at a large-scale. We expect our model to perform better in low resource scenarios because it can take advantage of the inherently hierarchical dialogue structure with induction bias. We conducted experiments over a range of training sizes (from 3k to 20k). As shown in Figure \ref{fig:low_resource}, our proposed \textit{PG-Net+TA} lead to steeper learning curves and always outperforms the baseline.

\begin{table}[t]
\begin{center}
\begin{tabular}{lccc}
\hline \textbf{Model} & \textbf{ROUGE-1} & \textbf{ROUGE-2} & \textbf{ROUGE-L}  \\ \hline
PG-Net & 42.33 & 28.23 & 38.14 \\
  & (-11.12) & (-14.88) & (-11.35) \\
PG-Net + TA & 49.70 & 39.32 & 45.37 \\
  & (-8.92) & (-9.74) & (-9.59) \\
\hline
\end{tabular}
\end{center}
\vspace{-0.3cm}
\caption{\label{lengthy-table}F1 scores from lengthy-sample evaluation. Bracketed values denote absolute decrease of model performance in Section \ref{ssec:effectiveness_eval}.}
\vspace{-0.2cm}
\end{table}

\subsection{Visualization Analysis}
In this section, we probe deeper into the proposed neural architecture to examine the innerworkings of (1) how the pointer and generator switches, and (2) how topic-level attention interacts with the decoded sequence. 

\subsubsection{Pointer-Generator Switching}
\label{sssec:switching_prob_vis}
We illustrate the probability of the pointer-generator switching $p_{gen}$ in Equation (\ref{eq:p-gen}) that indicates the probability of words generated from the vocabulary to show how the proposed model summarizes dialogue. One summary example produced from a dialogue with three topics is shown in Figure \ref{fig:attn_visual}a: the attribute information of a symptom and segment tags are directly copied from the source content, while symptom entities and punctuation tokens are generated from the vocabulary lexicon. This generation behavior resonates with the rationale used in constructing the ground-truth summary in Section \ref{ssec:groud-truth}, enabling the model to normalize symptom entities and handle out-of-vocabulary words in symptom attributes.

\subsubsection{Topic-Level Attention Scoring}
To show how the proposed framework conducts topic-aware contextual modeling, we illustrate topic-level attention scores $a^{seg}$ in Equation (\ref{eq:topic_attn}). As shown in Figure \ref{fig:attn_visual}b, during the summary decoding process of a dialogue with three topics, at each step, the model concentrates on one topic segment. We also observe smooth topic transition from the attention layer, which aligns well with the topical flow of the dialogue content. 
Such topic level modeling can help improve summarization performance by 
filtering out nonessential details at the word-level modeling layers.

\section{Conclusion}
\label{sec:conclusion}
In this work, we automatically summarized spoken dialogues from nurse-to-patient conversations. We presented an effective and efficient neural architecture that integrates topic-level attention mechanism in pointer-generator networks, utilizing the hierarchical structure of dialogues. We demonstrated that the proposed model significantly outperforms competitive baselines, obtains more efficient learning outcomes, is robust to lengthy dialogue samples, and performs well when there is limited training data.

\bibliographystyle{IEEEbib}
\bibliography{strings,refs}

\end{document}